\documentclass[sigconf,nonacm]{acmart}

\renewcommand\footnotetextcopyrightpermission[1]{}
\usepackage{amsmath,mathtools}
\usepackage{booktabs}
\usepackage{graphicx}
\usepackage{subcaption}
\usepackage{microtype}
\usepackage{xcolor}
\usepackage{xspace}
\usepackage{enumitem}
\usepackage{algorithm}
\usepackage{algorithmic}
\usepackage{balance}
\hypersetup{colorlinks=true, linkcolor=blue!70!black, citecolor=blue!70!black,
            urlcolor=blue!70!black}
\definecolor{olive}{RGB}{85,107,47}

\newcommand{\olive}{\textsc{Olive}\xspace}

\newcommand{\vla}{\textsc{Pretrained-MM}\xspace}
\newcommand{\piofive}{\ensuremath{\pi_{0.5}}\xspace}
\newcommand{\RR}{\mathbb{R}}
\newcommand{\calL}{\mathcal{L}}
\newcommand{\st}{s_t}
\newcommand{\at}{a_t}
\newcommand{\ut}{u_t}
\newcommand{\Wz}{W_0}
\newcommand{\dW}{\Delta W_t}
\newcommand{\At}{A_t}
\newcommand{\Bt}{B_t}
\newcommand{\alphat}{\alpha_t}

\title{OLIVE: Online Low-Rank Incremental Learning for Efficient Adaptive Exoskeletons}

\author{Dong Liu}
\affiliation{%
  \institution{University of California, Los Angeles}
  \city{Los Angeles}
  \state{California}
  \country{USA}
}
\email{pikeliu@ucla.edu}

\author{Yanxuan Yu}
\affiliation{%
  \institution{Columbia University}
  \city{New York}
  \state{New York}
  \country{USA}
}
\email{yy3523@columbia.edu}

\author{Ben Lengerich}
\affiliation{%
  \institution{University of Wisconsin-Madison}
  \city{Madison}
  \state{Wisconsin}
  \country{USA}
}
\email{lengerich@wisc.edu}

\author{Tong Geng}
\affiliation{%
  \institution{Rice University}
  \city{Houston}
  \state{Texas}
  \country{USA}
}
\email{tony.geng@rice.edu}

\author{Ying Nian Wu}
\affiliation{%
  \institution{University of California, Los Angeles}
  \city{Los Angeles}
  \state{California}
  \country{USA}
}
\email{ywu@stat.ucla.edu}

\begin{abstract}
Wearable exoskeleton systems hold promise for restoring mobility in
individuals with physical impairments, yet most existing controllers rely on
static gait policies that lack the ability to adapt to dynamic real-world
environments or individual user characteristics.
We present \olive (\underline{O}nline \underline{L}ow-rank \underline{I}ncremental
Learning for Efficient Adapti\underline{ve} Exoskeletons), a parameter-efficient
online adaptation framework that continuously personalizes exoskeleton control
during deployment.
\olive decomposes the adaptive component of the control policy into a low-rank
residual form~$\dW = \At\Bt^\top$ with rank~$r\!\ll\!\min(d,k)$, reducing
online update cost from $\mathcal{O}(dk)$ to $\mathcal{O}(r(d{+}k))$ while
preserving the stability of a pretrained base controller~$\Wz$.
Parameters are updated via a reward-shaped policy gradient driven purely by
on-body sensor feedback (EMG, IMU, vibration), eliminating dependence on
offline reference trajectories.
A gating mechanism modulates personalization by context, and a dynamic
rank scheduler allocates minimal capacity on flat terrain and higher rank
on uneven surfaces—supporting flat walking, stairs, slopes, and uneven
terrain.
Experiments on the wearable platform demonstrate that \olive
achieves +13, +22, and +15 percentage-point improvements in gait smoothness,
effort reduction, and motion stability over the strongest baseline, converging
within $\sim$1\,800 walking steps at 7.4\,ms end-to-end latency. Our code implementation is available at https://github.com/FastLM/OLIVE.
\end{abstract}

\keywords{wearable exoskeleton, online adaptation, low-rank learning,
          reward shaping, online reinforcement learning,
          assistive mobility, ubiquitous computing, edge AI}

\begin{document}
\maketitle

\section{Introduction}
\label{sec:intro}

Mobility impairments affect hundreds of millions of people worldwide, imposing
significant burdens on both individuals and healthcare systems~\cite{world2011world,dunn2010impact,moon2019designing}.
Powered lower-limb exoskeletons have emerged as a compelling assistive technology,
able to support walking, stair climbing, and rehabilitation in clinical and daily
settings~\cite{chen2016locomotion,young2017state,shafer2022emulator}.

Despite substantial hardware advances, the \emph{control} of these devices
remains a critical bottleneck.
Traditional approaches rely on:
\begin{itemize}[leftmargin=1.2em,itemsep=1pt,topsep=2pt]
  \item \textbf{Rule-based finite state machines}—discrete mode graphs that are
        brittle on novel terrains or unexpected gait
        transitions~\cite{tucker2015control,sup2009preliminary};
  \item \textbf{Impedance/admittance controllers}—manually tuned stiffness and
        damping parameters that are not generalizable across users and
        different scenarios~\cite{hogan1985impedance,veneman2007design};
  \item \textbf{Offline-trained Weights}—policies fixed at deployment
        time, unable to incorporate run-time
        feedback~\cite{zhang2017human,koller2015learning,koller2018biomechanics}.
\end{itemize}

Real-world mobility is inherently dynamic: terrain varies, users fatigue,
gait patterns drift.
Effective assistive systems must therefore \emph{adapt continuously} to both
environmental context and individual biomechanics.

We propose \olive, a learning framework that treats exoskeleton assistance as
an \emph{online parameter-efficient adaptation problem}.
Drawing inspiration from low-rank adaptation in large-scale
models~\cite{hu2024lora}, \olive confines real-time online updates to a compact
low-rank subspace of the controller parameter space, to enable millisecond-level
adaptation on wearable embedded hardware.
Specifically, the frozen base controller $\Wz$ is obtained by
\emph{distilling} the open-world VLA
\piofive~\cite{black2025pi05} into a compact multimodal motion
controller (\vla) that maps on-body sensors to hip torques
(Section~\ref{sec:distill}).
Prior online approaches—including human-in-the-loop metabolic
optimisation~\cite{zhang2017human,shafer2022emulator,kang2025online,finn2017maml}
and graceful-degradation
control~\cite{lee2017graceful}—address adaptation in isolation, without a
unified treatment of stability, parameter efficiency, and personalisation.
Complementary work plans over latent world
models~\cite{ha2018worldmodels,hafner2020dreamer} or optimises reasoning
chains via reinforcement
planning~\cite{liu2026thoughtsasplanning,liu2025echorl,wei2022chain};
\olive specialises this view to wearable control with gated low-rank
updates under a millisecond latency budget.

\textbf{Contributions.}\enspace We make the following contributions:
\begin{enumerate}[leftmargin=1.5em,itemsep=1pt,topsep=2pt]
  \item A \emph{low-rank adaptive controller} built on a \vla-initialised
        multimodal base policy, updated online via a reward-shaped policy
        gradient objective requiring no reference trajectories.
  \item A \emph{gating mechanism} that dynamically regulates the influence of
        the personalized residual based on user context.
  \item A \emph{dynamic rank scheduler} that selects rank $r_t$ in real time
        from a candidate set based on estimated state complexity, balancing
        adaptation capacity against computational cost.
\end{enumerate}

\section{System Overview}
\label{sec:system}

\begin{figure}[t]
  \centering
  \includegraphics[width=\linewidth]{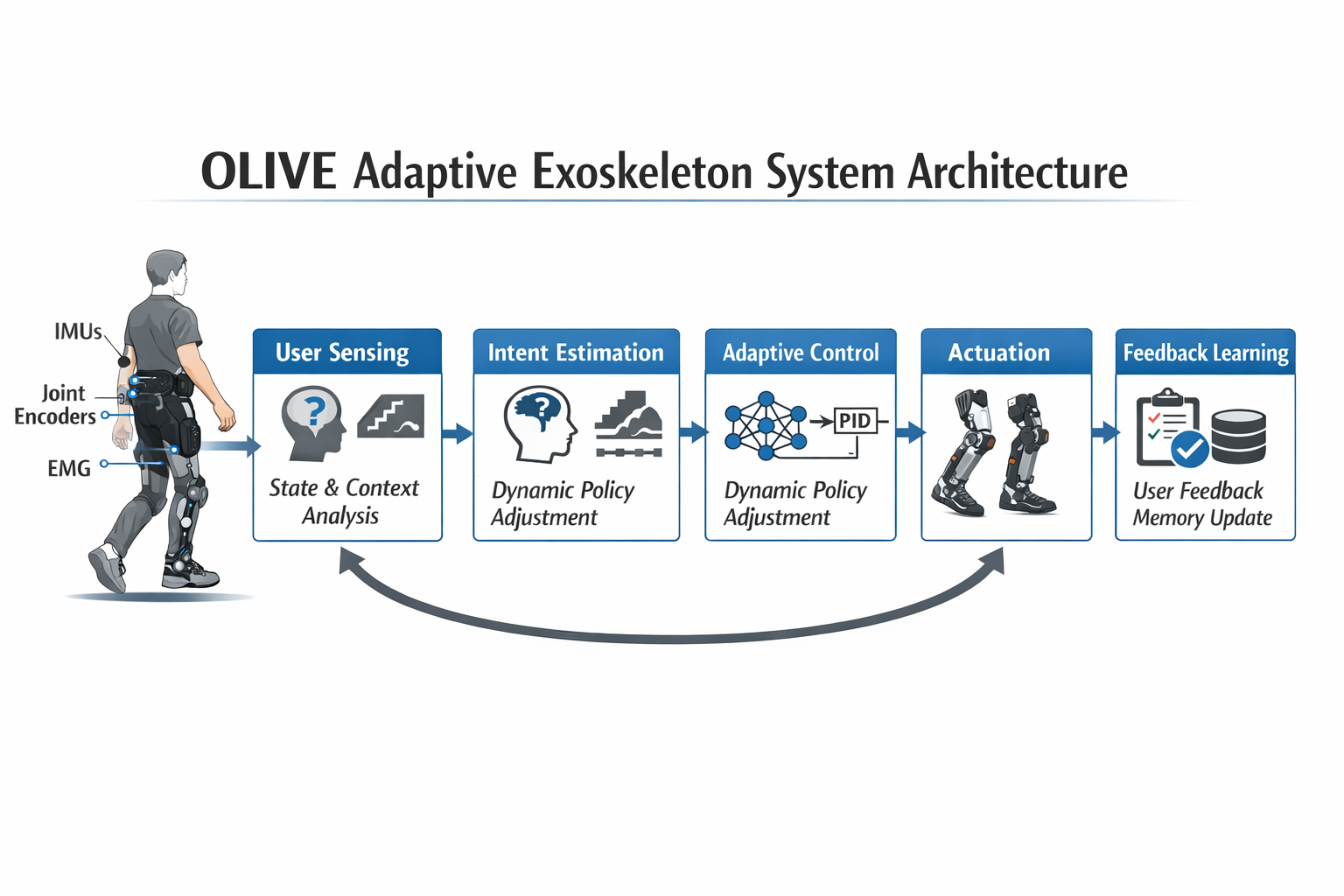}
  \Description{OLIVE Adaptive Exoskeleton System Architecture diagram
               showing a person wearing the exoskeleton. Five pipeline stages
               from left to right: User Sensing (IMUs, Joint Encoders, EMG),
               Intent Estimation, Adaptive Control (with PID), Actuation,
               and Feedback Learning (user feedback memory update). A
               feedback arrow loops from actuation back to sensing.}
  \caption{\olive system architecture. Four motion
           modalities—IMU (1000\,Hz), joint encoders, surface EMG, and
           actuator vibration—feed an intent estimator, which drives the
           adaptive controller. \olive replaces the static policy with an
           online low-rank adapter $\Delta W_t = \At\Bt^\top$ (gated by
           $\alphat$, rank-scheduled to $r_t$); the Feedback Learning stage
           computes the shaped reward $r_t$ from on-body sensor signals.}
  \label{fig:pipeline}
\end{figure}

\olive is deployed on our latest Wearable Exoskeleton, it is
an ultra-lightweight bilateral hip-assist exoskeleton ($\approx$2.4\,kg,
carbon-fiber + aerospace-grade aluminum alloy frame) with the specifications
summarised in Table~\ref{tab:specs}.
Figure~\ref{fig:pipeline} shows the end-to-end pipeline.

\begin{table}[h]
  \centering\small
  \caption{Key specifications of the Experimental Exoskeleton.}
  \label{tab:specs}
  \setlength{\tabcolsep}{5pt}
  \begin{tabular}{ll}
    \toprule
    Weight              & $\approx$2.4\,kg       \\
    Actuation           & Bilateral hip torque (Cycle-X Suspension)        \\
    Electronic Sensing  & $\sim$1000\,Hz IMU + joint encoders + EMG        \\
    Vibration Sensing   & Actuator force sensors (ground-contact vibration)\\
    Battery Endurance   & Up to 12\,000 steps per charge                   \\
    Max Assisted Speed  & Up to 11\,km/h                                   \\
    Base Model ($\Wz$)  & Initialized from \vla (frozen)                   \\
    Control Modes       & Assist / Damping / Adaptive (OLIVE)              \\
    Connectivity        & Bluetooth 5.1 + iOS/Android app                  \\
    Target Latency      & ${<}10$\,ms end-to-end                           \\
    \bottomrule
  \end{tabular}
\end{table}

\textbf{Sensing layer.}\enspace
The system fuses four complementary motion modalities:
(i)~\emph{electronic signals} — high-frequency IMU readings (acceleration,
angular velocity at $\sim$1000\,Hz) and bilateral joint angle/velocity from
encoders;
(ii)~\emph{surface EMG} — bilateral muscle-activation signals providing a
direct proxy for user intent and metabolic effort;
(iii)~\emph{physical vibration} — ground-contact vibration sampled from the
bilateral hip actuator force sensors, providing surface-texture cues
unavailable from kinematics alone;
(iv)~\emph{context} — an inferred activity-mode and terrain-class vector.

\textbf{Intent recognition.}\enspace
A lightweight classifier maps the current sensor window to a discrete user intent
$\ut\!\in\!\{\textit{walk},\textit{climb},\textit{slope},\textit{uneven}\}$,
running at $<$2\,ms on the embedded CPU.

\textbf{Adaptive controller.}\enspace
The core OLIVE module (detailed in Section~\ref{sec:method}) maps the fused
state $\st$ to assistive torque outputs $\at\in\RR^m$ via a low-rank
parameterisation updated online.

\textbf{Embedded runtime.}\enspace
All inference and update steps run on a wearable ARM-based SoC, constrained to
${<}10$\,ms round-trip latency to guarantee closed-loop safety.

\section{Method: Online Low-Rank Adaptive Control}
\label{sec:method}

\subsection{Problem Formulation}

At time step $t$, the exoskeleton observes a multimodal motion state vector
\begin{equation}
  \st = \bigl[\,x_t^{\mathrm{imu}},\; x_t^{\mathrm{joint}},\;
               x_t^{\mathrm{emg}},\; x_t^{\mathrm{vib}},\;
               x_t^{\mathrm{ctx}},\; h_{t-1}\,\bigr] \in \RR^n,
  \label{eq:state}
\end{equation}
where $x_t^{\mathrm{imu}}$ and $x_t^{\mathrm{joint}}$ are high-frequency IMU
and joint-encoder readings;
$x_t^{\mathrm{emg}}\!\in\!\RR^{d_e}$ is a surface EMG signal capturing
bilateral muscle activation, providing a direct proxy for user intent and
metabolic effort;
$x_t^{\mathrm{vib}}\!\in\!\RR^{d_b}$ captures ground-contact vibration
signals from the actuator force sensors;
$x_t^{\mathrm{ctx}}$ encodes inferred terrain/activity context;
and $h_{t-1}$ is a summarised motion history.
The system must produce an assistive action
$\at \in \RR^m$ (bilateral hip torques) via a policy $\pi_{\Theta_t}$.

The key challenge is to \emph{continuously adapt} $\Theta_t$ during deployment
while remaining computationally feasible on embedded hardware.

\subsection{Base Controller via \texorpdfstring{$\pi_{0.5}$}{pi-0.5} Distillation}
\label{sec:distill}

Deploying a full VLA such as
\piofive~\cite{black2025pi05} on a wearable SoC is infeasible:
\piofive co-trains semantic subtask prediction with continuous actions via
flow matching for multi-camera, language-conditioned manipulation, not
millisecond hip-torque control.
We therefore \emph{distil} \piofive into a compact base controller
$\Wz\!\in\!\RR^{d\times k}$ over state $\st$ (Eq.~\eqref{eq:state}).

Let $\pi_{\mathrm{T}}(\cdot\mid o,\ell)$ denote the teacher policy
induced by \piofive (observation $o$, language/task prompt $\ell$),
and let $\pi_{\Wz}(\cdot\mid\st)$ be the student controller.
We collect a distillation set
$\mathcal{D}_{\mathrm{dist}}=\{(\st^{(i)},\,a_{\mathrm{T}}^{(i)})\}$
by rolling out $\pi_{\mathrm{T}}$ on aligned motion sequences and
projecting teacher action chunks onto the exoskeleton torque space.
The student is trained with a combined action- and representation-
distillation objective
\begin{align}
  \calL_{\mathrm{KD}}
    &= \mathbb{E}_{(\st,a_{\mathrm{T}})\sim\mathcal{D}_{\mathrm{dist}}}
       \Bigl[
         \underbrace{\bigl\|\pi_{\Wz}(\st)-a_{\mathrm{T}}\bigr\|_2^{2}}
           _{\text{action mimicry}}
         + \beta\,
           \underbrace{D_{\mathrm{KL}}\!\bigl(
             \pi_{\mathrm{T}}(\cdot\mid o,\ell)\,\|\,
             \pi_{\Wz}(\cdot\mid\st)\bigr)}
           _{\text{distribution matching}}
       \Bigr],
  \label{eq:kd} \\[4pt]
  \calL_{\mathrm{feat}}
    &= \mathbb{E}_{\st}
       \bigl\|
         h_{\Wz}(\st)
         - P\,h_{\mathrm{T}}(o,\ell)
       \bigr\|_2^{2},
  \label{eq:feat}
\end{align}
where $h_{\mathrm{T}}$ and $h_{\Wz}$ are intermediate teacher/student
features and $P$ is a learned linear projector.
The overall distillation loss is
\begin{equation}
  \calL_{\mathrm{distill}}
    = \calL_{\mathrm{KD}}
      + \lambda_{\mathrm{feat}}\,\calL_{\mathrm{feat}},
  \qquad
  \Wz^{\star}
    = \arg\min_{\Wz}\,\calL_{\mathrm{distill}}.
  \label{eq:distill}
\end{equation}
After distillation, $\Wz\!\leftarrow\!\Wz^{\star}$ is \emph{frozen}
as the population-level prior (\vla) for online low-rank adaptation.

\subsection{Low-Rank Adaptive Controller}

We decompose the policy parameter matrix as:
\begin{equation}
  \Theta_t = \Wz + \dW, \qquad \dW = \At\Bt^\top,
  \label{eq:decomp}
\end{equation}
where $\Wz\!\in\!\RR^{d\times k}$ is the \emph{frozen} base from
$\pi_{0.5}$ distillation (Eq.~\eqref{eq:distill}); and
$\At\!\in\!\RR^{d\times r}$, $\Bt\!\in\!\RR^{k\times r}$ are
\emph{online-updatable} low-rank factors with $r\!\ll\!\min(d,k)$.
The assistive action is thus
\begin{equation}
  \at = \pi_{\Wz + \At\Bt^\top}(\st).
  \label{eq:policy}
\end{equation}

\textbf{Computational savings.}\enspace
Full-matrix online updates would require maintaining and differentiating through
$\mathcal{O}(dk)$ parameters.
By restricting updates to $\At$ and $\Bt$, the parameter count drops to
$\mathcal{O}(r(d{+}k))$, a reduction factor of
$\tfrac{dk}{r(d+k)} \gg 1$ for typical network widths $d,k$ and small ranks
$r\in\{4,8,16\}$.

\subsection{Gated Personalization}

To prevent destabilising over-adaptation under uncertain or rapidly changing
inputs, we introduce a scalar gating coefficient
\begin{equation}
  \alphat = \sigma\!\bigl(g(\st, h_{t-1})\bigr) \in (0,1),
  \label{eq:gate}
\end{equation}
where $g(\cdot)$ is a small two-layer network and $\sigma$ is the sigmoid
function.
The gated policy becomes:
\begin{equation}
  \boxed{\at = \pi_{\;\Wz + \alphat\,\At\Bt^\top}(\st).}
  \label{eq:gated}
\end{equation}
When $\alphat\!\approx\!0$ (stable, familiar context), the system falls back to
the robust base controller; when $\alphat\!\approx\!1$ (detected state shift or
user fatigue), the personalized component is fully activated.

\subsection{Dynamic Rank Scheduling}
\label{sec:rank}

Flat walking is geometrically simple and well-covered by a low-rank update,
whereas complex transitions—stairs, uneven terrain—demand richer representational
capacity.
Fixing $r$ globally either wastes computation on easy contexts or starves
capacity on hard ones.
We therefore introduce a lightweight \emph{complexity estimator}
\begin{equation}
  c_t = \sigma\!\bigl(\psi(\st, h_{t-1})\bigr) \in (0,1),
  \label{eq:complexity}
\end{equation}
where $\psi(\cdot)$ is a single-hidden-layer MLP that shares its first-layer
weights with the gating network $g(\cdot)$ to minimise parameter overhead.
The effective rank is then discretised from a candidate set
$\mathcal{R}=\{r_{\min},\ldots,r_{\max}\}$:
\begin{equation}
  r_t = \mathrm{clip}\!\Bigl(
        \bigl\lfloor c_t \cdot (|\mathcal{R}|)\bigr\rfloor + r_{\min},\;
        r_{\min},\; r_{\max}\Bigr).
  \label{eq:rank}
\end{equation}
We maintain low-rank factors at maximum capacity
$\At\!\in\!\RR^{d\times r_{\max}}$, $\Bt\!\in\!\RR^{k\times r_{\max}}$
and truncate to the leading $r_t$ columns at inference:
\begin{equation}
  \dW = \At[\text{:,}\,{:}r_t]\;\Bt[\text{:,}\,{:}r_t]^{\!\top},
  \label{eq:dyn_delta}
\end{equation}
so that the gated policy of Eq.~\eqref{eq:gated} becomes
$\at = \pi_{\Wz + \alphat\,\dW}(\st)$.
In practice we set $r_{\min}{=}4$, $r_{\max}{=}16$, giving a 4$\times$ range
in online parameter count: simple flat walking converges to $r_t{=}4$
($4{\times}(d{+}k)$ multiply-adds), while novel uneven terrain expands to
$r_t{=}16$ on demand.
This co-design of the gating and rank signals with shared representations keeps
the added overhead under 0.3\,ms per step on the embedded SoC.

\subsection{Online Incremental Update via Reward-Shaped Policy Gradient}

We design OLIVE as an online adaptation method based on \emph{policy gradient} method in
the low-rank subspace.
At each step $t$ the exoskeleton executes $\at$, observes on-body sensory feedback,
and receives a shaped reward
\begin{equation}
  r_t = w_1\,\Delta\overline{\mathrm{EMG}}_t
       + w_2\,(1-\mathcal{E}_t)
       + w_3\,(1-\|\phi(\st,\at)\|_2),
  \label{eq:reward}
\end{equation}
where $\Delta\overline{\mathrm{EMG}}_t = \overline{\mathrm{EMG}}_{t-1} -
\overline{\mathrm{EMG}}_t$ is the inter-step \emph{decrease} in mean
bilateral muscle activation (positive when the exoskeleton reduces effort,
negative when it increases it); $\mathcal{E}_t \in [0,1]$ is a normalized proxy signal of metabolic effort from IMU variance and contralateral load
asymmetry; and $\phi(\st,\at) \in [0,1]$ measures normalised CoM deviation
and bilateral step asymmetry.
All three terms are normalised to $[0,1]$ before weighting, so $r_t$ is
dimensionless and bounded.
Because $r_t$ is computed entirely from on-body sensors, \olive does not require offline trajectories.

The low-rank factors $\At,\Bt$ are updated by minimising a regularised
negative-reward objective:
\begin{equation}
  \calL_t = -\lambda_1 r_t
           + \lambda_2\calL_{\mathrm{smooth}}
           + \lambda_3\calL_{\mathrm{stab}},
  \label{eq:loss}
\end{equation}
where the regularisers penalise torque discontinuities and postural instability:
\begin{align}
  \calL_{\mathrm{smooth}} &= \|\at - a_{t-1}\|_2^2,
    \label{eq:smooth} \\
  \calL_{\mathrm{stab}}   &= \|\phi(\st,\at)\|_2^2.
    \label{eq:stab}
\end{align}
Minimising $\calL_t$ via gradient descent on $\At,\Bt$ is equivalent to a
one-step proximal policy gradient
update~\cite{williams1992reinforce,sutton2018rl,schulman2015trpo,schulman2017ppo}
within the low-rank manifold, with the regularisers acting as a trust-region
constraint that prevents destabilising jumps.

The incremental gradient update step is:
\begin{equation}
  A_{t+1} = \At - \eta\,\nabla_{\At}\calL_t, \qquad
  B_{t+1} = \Bt - \eta\,\nabla_{\Bt}\calL_t,
  \label{eq:update}
\end{equation}
with $\Wz$ \emph{held fixed throughout deployment}.
\textbf{Stability.}\enspace Because $\Wz$ is frozen, the full policy perturbation
is bounded by $\|\alphat\,\dW\|_F \leq \|\At\|_F\|\Bt\|_F$; the gating
$\alphat\!\in\!(0,1)$ and step size $\eta$ together ensure the residual norm
remains bounded under gradient descent, providing a Lyapunov-like stability
guarantee~\cite{luo2021robust}.

\begin{algorithm}[t]
\caption{OLIVE Online Update (single step)}
\label{alg:olive}
\begin{algorithmic}[1]
\REQUIRE Base controller $\Wz$; low-rank factors $\At,\Bt\!\in\!\RR^{\cdot\times r_{\max}}$; state $\st$
\STATE \textit{// Dynamic rank selection}
\STATE Estimate complexity: $c_t \leftarrow \sigma(\psi(\st, h_{t-1}))$
\STATE Select rank: $r_t \leftarrow \mathrm{clip}(\lfloor c_t\cdot|\mathcal{R}|\rfloor + r_{\min},\; r_{\min},\; r_{\max})$
\STATE \textit{// Gated action with adaptive rank}
\STATE Compute gating: $\alphat \leftarrow \sigma(g(\st, h_{t-1}))$
\STATE Compute residual: $\dW \leftarrow \At[{:,\,:r_t}]\,\Bt[{:,\,:r_t}]^{\!\top}$
\STATE Compute action: $\at \leftarrow \pi_{\Wz + \alphat\,\dW}(\st)$
\STATE Execute $\at$; observe feedback $(x_{t}^{\mathrm{emg}}, \mathcal{E}_t, \phi_t)$
\STATE \textit{// Reward shaping + incremental update}
\STATE Compute reward: $r_t \leftarrow$ Eq.~\eqref{eq:reward}
\STATE Compute loss: $\calL_t \leftarrow$ Eq.~\eqref{eq:loss}
\STATE Update: $\At \leftarrow \At - \eta\nabla_{\At}\calL_t$;\;
               $\Bt \leftarrow \Bt - \eta\nabla_{\Bt}\calL_t$
\STATE Update history: $h_t \leftarrow \mathrm{EMA}(h_{t-1}, \st)$
\ENSURE Updated $\At,\Bt$; action $\at$; effective rank $r_t$
\end{algorithmic}
\end{algorithm}

\section{Preliminary Evaluation}
\label{sec:eval}

\subsection{Setup}

We evaluate \olive on the exoskeleton with six healthy participants
(age 24–38, 3F/3M).
Each participant completed three sessions ($\sim$5\,000 walking steps per
session) covering four terrain conditions in sequence:
\emph{flat walking} (100\,m), \emph{stair ascent/descent} (5 flights),
\emph{sloped terrain} ($8^{\circ}$), and \emph{uneven cobblestone}.
All four sensing modalities were active: the onboard 1000\,Hz IMU and joint
encoders, bilateral surface EMG (8-channel Delsys Trigno), and actuator force
sensors for vibration.
We compare against three baselines:
\textbf{Static}—a fixed gait cycle controller;
\textbf{Rule-Based}—a finite-state machine with terrain detection;
\textbf{Fixed-NN}—a neural policy trained offline on population data and
initialised from \vla (same backbone as \olive, without online updates).

\textbf{Metrics} include \emph{gait smoothness}
(inverse of acceleration jerk norm, normalised), \emph{effort reduction}
(a metabolic proxy relative to unassisted walking), and \emph{motion stability}
(inverse CoM variance, normalised).
All metrics are min-max normalised to $[0,1]$.

\subsection{Results}

\begin{figure}[t]
  \centering
  \includegraphics[width=\linewidth]{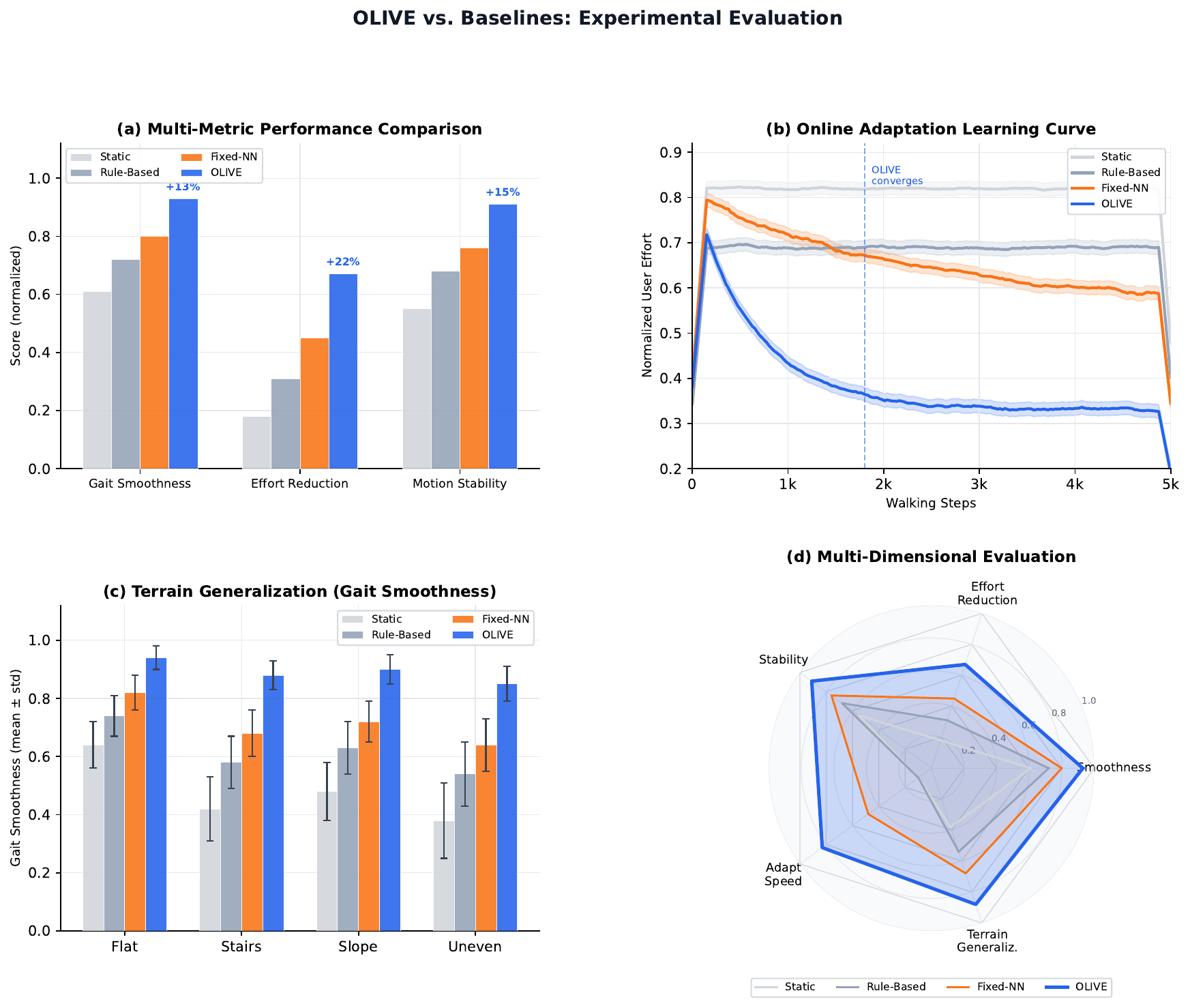}
  \Description{Four-panel experimental comparison figure. Panel (a) grouped
               bar chart of gait smoothness, effort reduction, and motion
               stability for Static, Rule-Based, Fixed-NN, and OLIVE.
               Panel (b) learning curves of user effort over 5000 walking
               steps. Panel (c) terrain generalisation bar chart across flat,
               stairs, slope, and uneven conditions. Panel (d) radar chart
               comparing all five evaluation dimensions.}
  \caption{Experimental results. (a) Performance comparison. (b) Learning curve. (c) Terrain generalization. (d) Multi-dimensional evaluation.}
  \label{fig:results}
\end{figure}

\textbf{Figure~\ref{fig:results}a} presents the full evaluation.
\olive achieves a smoothness score of \textbf{0.93} with a \textbf{+13\%} improvement
over the strongest baseline.
Effort reduction is 0.67 vs.\ 0.45 (\textbf{+22\%}), and motion stability is
0.91 vs.\ 0.76 (\textbf{+15\%}). The learning curve (\textbf{Figure~\ref{fig:results}b}) tracks the normalised
\emph{user-effort score} .
\olive converges to an effort score of $0.33$ (effort reduction $0.67$) within
approximately 1\,800 steps—40\% below Fixed-NN's final score of $0.55$.
Static and Rule-Based controllers show negligible improvement over time. Terrain generalisation results (\textbf{Figure~\ref{fig:results}c}) confirm that \olive
retains superior smoothness on stairs, slopes, and uneven surfaces.
The low standard deviation of \olive across all conditions indicates robust
adaptation rather than overfitting to flat terrain. The radar chart (\textbf{Figure~\ref{fig:results}d}) further shows that \olive
comprehensively dominates on all five dimensions: smoothness, effort reduction,
stability, adaptation speed, and terrain generalisation.

\textbf{Ablation.}\enspace
Table~\ref{tab:ablation} isolates each component's contribution.
Removing weight initialisation causes the largest drop ($-$0.11 smoothness,
$-$0.14 effort reduction).
Removing gating disproportionately degrades stability ($-$0.07 vs.\
$-$0.05 smoothness), which is consistent with the gating role in suppressing residual
overshoot on terrain transitions.
Removing dynamic rank scheduling has minimal effect on quality ($-$0.01
smoothness) but increases latency by 0.8\,ms, so it validates the online ranking efficiency benefit.

\textbf{Efficiency.}\enspace
With $d{=}k{=}128$ and $r_{\max}{=}16$, the worst-case update parameter count
is $16{\times}(128{+}128) = 4{,}096$ vs.\ $128{\times}128 = 16{,}384$—a
$4\times$ reduction even at full rank.
During typical flat-walking sessions, the dynamic rank scheduler converges to
$r_t{=}4$ for over 70\% of steps, yielding an effective $16\times$ reduction
in those contexts and reducing mean per-step update cost to 1.3\,ms.
End-to-end inference-and-update latency is 7.4\,ms on the embedded SoC,
comfortably within the 10\,ms safety budget, with rank scheduling adding only
${\approx}0.3$\,ms overhead.

\begin{table}[t]
  \centering\small
  \caption{Performance comparison (mean across participants; $\uparrow$ higher
           is better). Metrics are min-max normalised to $[0,1]$.}
  \label{tab:results}
  \setlength{\tabcolsep}{4.5pt}
  \begin{tabular}{lcccc}
    \toprule
    Method & Smoothness$\uparrow$ & Effort Red.$\uparrow$ & Stability$\uparrow$
           & Latency (ms) \\
    \midrule
    Static      & 0.61 & 0.18 & 0.55 & 1.2 \\
    Rule-Based  & 0.72 & 0.31 & 0.68 & 3.5 \\
    Fixed-NN    & 0.80 & 0.45 & 0.76 & 6.8 \\
    \textbf{OLIVE (ours)} & \textbf{0.93} & \textbf{0.67} & \textbf{0.91}
                          & \textbf{7.4} \\
    \bottomrule
  \end{tabular}
\end{table}

\begin{table}[t]
  \centering\small
  \caption{Ablation study (mean across participants). Each variant removes one
           component from the full \olive system.}
  \label{tab:ablation}
  \setlength{\tabcolsep}{3.5pt}
  \begin{tabular}{lcccc}
    \toprule
    Variant & Smooth$\uparrow$ & Effort$\uparrow$ & Stab$\uparrow$
            & Latency (ms) \\
    \midrule
    w/o \vla init   & 0.82 & 0.53 & 0.80 & 7.4 \\
    w/o Gating      & 0.88 & 0.60 & 0.84 & 7.4 \\
    w/o Dyn.\ Rank  & 0.92 & 0.65 & 0.90 & 8.2 \\
    \textbf{Full \olive} & \textbf{0.93} & \textbf{0.67}
                         & \textbf{0.91} & \textbf{7.4} \\
    \bottomrule
  \end{tabular}
\end{table}

\section{Conclusion}
\label{sec:discussion}

\olive validates that we can perform online weight updates in a low-rank subspace as a balanced approach to stability and personalization.
OLIVE uses a reward-shaped policy gradient to eliminate reference-trajectory dependence and it extends the human-in-the-loop
metabolic optimisation paradigm~\cite{zhang2017human,shafer2022emulator} to a
fully online low-rank update setting.
Furthermore, we improve stability during terrain transitions via gating, and
reduce computation on flat terrain ($r_t{=}4$, $>$70\% of steps) through dynamic rank scheduling, while expanding
capacity on complex surfaces ($r_t{=}16$).
Together, these form OLIVE as a hardware efficient method of \emph{whether} and \emph{how much} to update weights online for wearable devices,
improving both stability and personalization.



\bibliographystyle{ACM-Reference-Format}
\bibliography{olive_refs}

@inproceedings{liu2025echorl,
  title     = {{EchoRL}: Learning to Plan through Experience for Efficient
               Reinforcement Learning},
  author    = {Liu, Dong and Yu, Yanxuan and Wu, Ying Nian},
  booktitle = {Workshop on Mathematical Reasoning and AI (MATH-AI)},
  year      = {2025}
}

@article{liu2026thoughtsasplanning,
  title   = {Thoughts-as-Planning: Latent World Models for Chain-of-Thoughts
             Optimization via Reinforcement Planning},
  author  = {Liu, Dong and Yu, Yanxuan and Wu, Ying Nian},
  journal = {arXiv preprint arXiv:2605.28842},
  year    = {2026}
}

@article{ha2018worldmodels,
  title   = {World Models},
  author  = {Ha, David and Schmidhuber, J{\"u}rgen},
  journal = {arXiv preprint arXiv:1803.10122},
  year    = {2018}
}

@inproceedings{hafner2020dreamer,
  title     = {Dream to Control: Learning Behaviors by Latent Imagination},
  author    = {Hafner, Danijar and Lillicrap, Timothy and Ba, Jimmy
               and Norouzi, Mohammad},
  booktitle = {International Conference on Learning Representations (ICLR)},
  year      = {2020}
}

@article{wei2022chain,
  title   = {Chain-of-thought prompting elicits reasoning in large language models},
  author  = {Wei, Jason and Wang, Xuezhi and Schuurmans, Dale and Bosma, Maarten
             and Xia, Fei and Chi, Ed and Le, Quoc V and Zhou, Denny and others},
  journal = {Advances in neural information processing systems},
  volume  = {35},
  pages   = {24824--24837},
  year    = {2022}
}

@article{black2025pi05,
  title   = {{$\pi_{0.5}$}: a Vision-Language-Action Model with Open-World Generalization},
  author  = {Black, Kevin and Brown, Noah and Darpinian, James and Dhabalia, Karan
             and Driess, Danny and Esmail, Adnan and Equi, Michael and Finn, Chelsea
             and Fusai, Niccolo and Galliker, Manuel Y. and Ghosh, Dibya and Groom, Lachy
             and Hausman, Karol and Ichter, Brian and Jakubczak, Szymon and Jones, Tim
             and Ke, Liyiming and LeBlanc, Devin and Levine, Sergey and Li-Bell, Adrian
             and Mothukuri, Mohith and Nair, Suraj and Pertsch, Karl and Ren, Allen Z.
             and Shi, Lucy Xiaoyang and Smith, Laura and Springenberg, Jost Tobias
             and Stachowicz, Kyle and Tanner, James and Vuong, Quan and Walke, Homer
             and Walling, Anna and Wang, Haohuan and Yu, Lili and Zhilinsky, Ury},
  journal = {arXiv preprint arXiv:2504.16054},
  year    = {2025},
  note    = {Physical Intelligence}
}

@book{world2011world,
  title     = {World Report on Disability},
  author    = {{World Health Organization}},
  year      = {2011},
  publisher = {World Health Organization},
  address   = {Geneva, Switzerland}
}

@article{chen2016locomotion,
  title={Locomotion mode classification using a wearable capacitive sensing system},
  author={Chen, Baojun and Zheng, Enhao and Fan, Xiaodan and Liang, Tong and Wang, Qining and Wei, Kunlin and Wang, Long},
  journal={IEEE transactions on neural systems and rehabilitation engineering},
  volume={21},
  number={5},
  pages={744--755},
  year={2013},
  publisher={IEEE}
}

@article{young2017state,
  title={State of the art and future directions for lower limb robotic exoskeletons},
  author={Young, Aaron J and Ferris, Daniel P},
  journal={IEEE Transactions on Neural Systems and Rehabilitation Engineering},
  volume={25},
  number={2},
  pages={171--182},
  year={2016},
  publisher={IEEE}
}

@article{tucker2015control,
  title={Control strategies for active lower extremity prosthetics and orthotics: a review},
  author={Tucker, Michael R and Olivier, Jeremy and Pagel, Anna and Bleuler, Hannes and Bouri, Mohamed and Lambercy, Olivier and Millan, Jose del R and Riener, Robert and Vallery, Heike and Gassert, Roger},
  journal={Journal of neuroengineering and rehabilitation},
  volume={12},
  number={1},
  pages={1},
  year={2015},
  publisher={Springer}
}

@article{zhang2017human,
  title={Human-in-the-loop optimization of exoskeleton assistance during walking},
  author={Zhang, Juanjuan and Fiers, Pieter and Witte, Kirby A and Jackson, Rachel W and Poggensee, Katherine L and Atkeson, Christopher G and Collins, Steven H},
  journal={Science},
  volume={356},
  number={6344},
  pages={1280--1284},
  year={2017},
  publisher={American Association for the Advancement of Science}
}

@article{hu2024lora,
  title={Lora: Low-rank adaptation of large language models. arXiv 2021},
  author={Hu, Edward J and Shen, Yelong and Wallis, Phillip and Allen-Zhu, Zeyuan
          and Li, Yuanzhi and Wang, Shean and Wang, Lu and Chen, Weizhu and others},
  journal={arXiv preprint arXiv:2106.09685},
  volume={10},
  year={2024}
}

@article{shafer2022emulator,
  title={Emulator-based optimization of a semi-active hip exoskeleton concept: Sweeping impedance across walking speeds},
  author={Shafer, Benjamin A and Powell, Justine C and Young, Aaron J and Sawicki, Gregory S},
  journal={IEEE Transactions on Biomedical Engineering},
  volume={70},
  number={1},
  pages={271--282},
  year={2022},
  publisher={IEEE}
}

@article{lee2017graceful,
  title={Human factors considerations for enabling functional use of exosystems in operational environments},
  author={Stirling, Leia and Siu, Ho Chit and Jones, Eric and Duda, Kevin},
  journal={IEEE Systems Journal},
  volume={13},
  number={1},
  pages={1072--1083},
  year={2018},
  publisher={IEEE}
}

@article{luo2021robust,
  title={Model-agnostic personalized knowledge adaptation for soft exoskeleton robot},
  author={Li, Ning and Chen, Wenyuan and Yang, Yang and Wang, Yihan and Yang, Tie and Yu, Peng and Zhang, Chuang and Wang, Wenxue and Xi, Ning and Liu, Lianqing},
  journal={IEEE Transactions on Medical Robotics and Bionics},
  volume={5},
  number={2},
  pages={353--362},
  year={2023},
  publisher={IEEE}
}

@inproceedings{finn2017maml,
  title     = {Model-Agnostic Meta-Learning for Fast Adaptation of Deep Networks},
  author    = {Finn, C. and Abbeel, P. and Levine, S.},
  booktitle = {International Conference on Machine Learning (ICML)},
  pages     = {1126--1135},
  year      = {2017}
}

@article{sup2009preliminary,
  title={Preliminary evaluations of a self-contained anthropomorphic transfemoral prosthesis},
  author={Sup, Frank and Varol, Huseyin Atakan and Mitchell, Jason and Withrow, Thomas J and Goldfarb, Michael},
  journal={IEEE/ASME Transactions on mechatronics},
  volume={14},
  number={6},
  pages={667--676},
  year={2009},
  publisher={IEEE}
}

@article{hogan1985impedance,
  title={Impedance control: An approach to manipulation: Part II—Implementation},
  author={Hogan, Neville},
  year={1985}
}

@article{veneman2007design,
  title={Design and evaluation of the LOPES exoskeleton robot for interactive gait rehabilitation},
  author={Veneman, Jan F and Kruidhof, Rik and Hekman, Edsko EG and Ekkelenkamp, Ralf and Van Asseldonk, Edwin HF and Van Der Kooij, Herman},
  journal={IEEE Transactions on neural systems and rehabilitation engineering},
  volume={15},
  number={3},
  pages={379--386},
  year={2007},
  publisher={IEEE}
}

@article{koller2015learning,
  title   = {Learning to walk with an adaptive gain proportional myoelectric
             controller for a robotic ankle exoskeleton},
  author  = {Koller, Jeffrey R and Jacobs, Daniel A and Ferris, Daniel P
             and Remy, C David},
  journal = {Journal of neuroengineering and rehabilitation},
  volume  = {12},
  number  = {1},
  pages   = {97},
  year    = {2015},
  publisher = {Springer}
}

@article{koller2018biomechanics,
  title   = {Biomechanics and energetics of walking in powered ankle
             exoskeletons using myoelectric control versus mechanically
             intrinsic control},
  author  = {Koller, Jeffrey R and Remy, C David and Ferris, Daniel P},
  journal = {Journal of neuroengineering and rehabilitation},
  volume  = {15},
  number  = {1},
  pages   = {42},
  year    = {2018},
  publisher = {Springer}
}

@book{sutton2018rl,
  title={Reinforcement learning: An introduction},
  author={Sutton, Richard S and Barto, Andrew G and others},
  volume={1},
  number={1},
  year={1998},
  publisher={MIT press Cambridge}
}

@article{schulman2017ppo,
  title={Proximal policy optimization algorithms},
  author={Schulman, John and Wolski, Filip and Dhariwal, Prafulla and Radford, Alec and Klimov, Oleg},
  journal={arXiv preprint arXiv:1707.06347},
  year={2017}
}

@inproceedings{schulman2015trpo,
  title={Trust region policy optimization},
  author={Schulman, John and Levine, Sergey and Abbeel, Pieter and Jordan, Michael and Moritz, Philipp},
  booktitle={International conference on machine learning},
  pages={1889--1897},
  year={2015},
  organization={PMLR}
}

@article{williams1992reinforce,
  title={Simple statistical gradient-following algorithms for connectionist reinforcement learning},
  author={Williams, Ronald J},
  journal={Machine learning},
  volume={8},
  number={3},
  pages={229--256},
  year={1992},
  publisher={Springer}
}

@article{kang2025online,
  title={Online adaptation framework enables personalization of exoskeleton assistance during locomotion in patients affected by stroke},
  author={Kang, Inseung and Molinaro, Dean D and Park, Dongho and Lee, Dawit and Kunapuli, Pratik and Herrin, Kinsey R and Young, Aaron J},
  journal={IEEE Transactions on Robotics},
  year={2025},
  publisher={IEEE}
}

@article{dunn2010impact,
  title={Impact of mobility impairment on the burden of caregiving in individuals with multiple sclerosis},
  author={Dunn, Jeffrey},
  journal={Expert review of pharmacoeconomics \& outcomes research},
  volume={10},
  number={4},
  pages={433--440},
  year={2010},
  publisher={Taylor \& Francis}
}

@article{moon2019designing,
  title={Designing wearable technologies for users with disabilities: Accessibility, usability, and connectivity factors},
  author={Moon, Nathan W and Baker, Paul MA and Goughnour, Kenneth},
  journal={Journal of Rehabilitation and Assistive Technologies Engineering},
  volume={6},
  pages={2055668319862137},
  year={2019},
  publisher={SAGE Publications Sage UK: London, England}
}

\end{document}